\documentclass[a4paper, 10pt, conference]{ieeeconf}      
\IEEEoverridecommandlockouts                              
\makeatletter
\let\NAT@parse\undefined
\makeatother

\usepackage[dvipsnames]{xcolor}
\usepackage{balance}
\usepackage{flushend}
\usepackage{amsmath}

\newcommand*\linkcolours{ForestGreen}

\usepackage{times}
\usepackage{graphicx}
\graphicspath{
    {imgs/}
}
\usepackage{amssymb}
\usepackage{gensymb}
\usepackage{amsmath}
\usepackage{breakurl}

\PassOptionsToPackage{hyphens}{url}
\usepackage{hyperref}
\hypersetup{
colorlinks,
linkcolor=\linkcolours,
citecolor=\linkcolours,
filecolor=\linkcolours,
urlcolor=\linkcolours}

\usepackage{algorithm}
\usepackage{algorithmic}

\usepackage[labelfont={bf},font=small]{caption}
\usepackage[none]{hyphenat}

\usepackage{mathtools, cuted}

\usepackage{cleveref}

\usepackage{tabularx}
\usepackage{amsmath}
\usepackage[official]{eurosym}

\usepackage{float}

\usepackage{pifont}

\newcolumntype{Y}{>{\centering\arraybackslash}X}

\usepackage[]{placeins}

\usepackage{placeins}

\usepackage{tikz}

\usepackage[framemethod=tikz]{mdframed}

\usepackage{afterpage}

\usepackage{stfloats}

\usepackage{atbegshi}
\newcommand{\handlethispage}{}
\newcommand{\discardpagesfromhere}{\let\handlethispage\AtBeginShipoutDiscard}
\newcommand{\keeppagesfromhere}{\let\handlethispage\relax}
\AtBeginShipout{\handlethispage}
\usepackage{comment}

\author{
  Marc Tonsen \\
   \textbf{Pupil Labs} \\
   Germany, Berlin \\
   \texttt{mar@pupil-labs.com}
  \and
 Chris Kay Baumann \\
   \textbf{Pupil Labs}\\
   Germany, Berlin \\
   \texttt{ckb@pupil-labs.com} 
     \and
 Kai Dierkes \\
 \textbf{Pupil Labs} \\
 Germany, Berlin \\
 \texttt{kai@pupil-labs.com}
}

\title{{\centering \LARGE \bf \hspace*{2.2cm} A High-Level Description and}\newline\LARGE \bf Performance Evaluation of Pupil Invisible} 

\begin{document}

\maketitle
\thispagestyle{empty}
\pagestyle{empty}

\begin{abstract}
	
	Head-mounted eye trackers promise convenient access to reliable gaze data in unconstrained environments. 
	Due to several limitations, however, often they can only partially deliver on this promise. 
	
	Among those are the following: (i) the necessity of performing a device setup and calibration prior to every use of the eye tracker, (ii) a lack of robustness of gaze-estimation results against perturbations, such as outdoor lighting conditions and unavoidable slippage of the eye tracker on the head of the subject, and (iii) behavioral distortion resulting from social awkwardness, due to the unnatural appearance of current head-mounted eye trackers.
	
	Recently, Pupil Labs released Pupil Invisible glasses, a head-mounted eye tracker engineered to tackle these limitations. 
	Here, we present an extensive evaluation of its gaze-estimation capabilities.
	To this end, we designed a data-collection protocol and evaluation scheme geared towards providing a faithful portrayal of the real-world usage of Pupil Invisible glasses. 
	
	In particular, we develop a geometric framework for gauging gaze-estimation accuracy that goes beyond reporting mean angular accuracy.
	We demonstrate that Pupil Invisible glasses, without the need of a calibration, provide gaze estimates which are robust to perturbations, including outdoor lighting conditions and slippage of the headset.
	
\end{abstract}

\section{Introduction}
\label{sec:introduction}
Eye tracking has become an established tool enabling an ever expanding range of diverse applications \cite{KleEtt19}. 
Gaze and other eye-related data empower researchers to gain insights into human cognition \cite{RahFie19} and behavior \cite{MeiOll19}.
It is employed as a diagnostic tool for a number of medical conditions \cite{HarKas18}. 
Industrial applications include product design evaluation \cite{KhaGre15} and marketing studies \cite{SanOli15, RotRei19}. 
Much effort is put into devising novel human-computer-interaction patterns that use gaze as an input modality \cite{MajBul14}. 
\begin{figure}[t!]
	\centering
	\includegraphics[width=0.9\columnwidth]{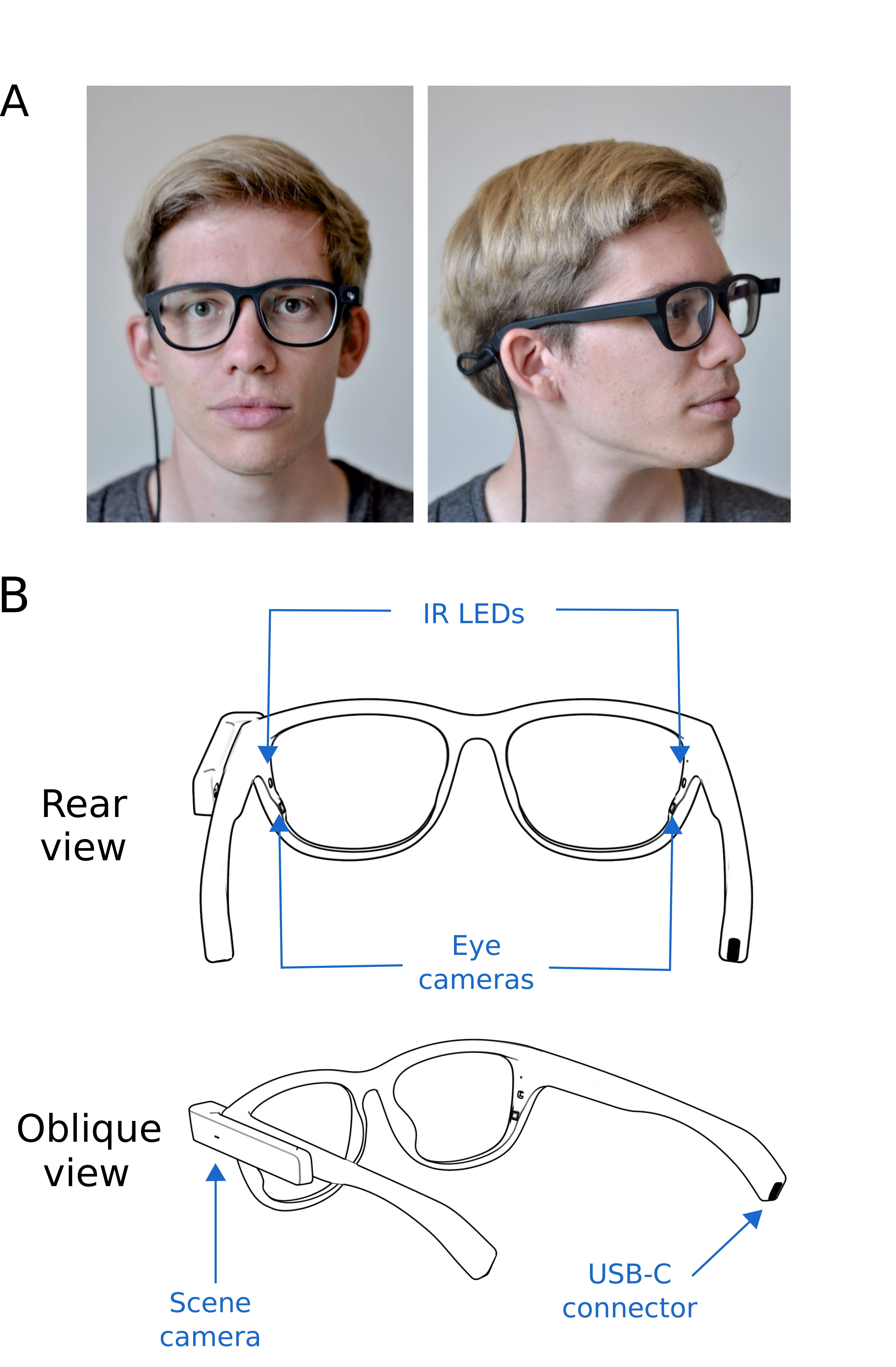}
	\caption{Components of Pupil Invisible glasses. (A) The hardware form factor of Pupil Invisible glasses is close to regular eye glasses and is thus expected to reduce social distortion. (B) Line drawings of Pupil Invisible glasses in a rear (top drawing) and oblique view (bottom drawing). Two eye cameras are embedded temporally into the glasses frame. IR LEDs in close proximity to the eye cameras are used for active illumination, making it possible to use Pupil Invisible glasses in dark environments. A wide-angle detachable scene camera module allows for recording the scene in front of the wearer. A USB-C connector is embedded into the right temple. In order to operate Pupil Invisible glasses, they need to be connected to a smart-phone running the Pupil Invisible Companion app.}
	\label{fig:component_overview}
\end{figure}

Over the last decades, a plethora of technological solutions to the eye-tracking challenge has been proposed \cite{HanQia10, Hut19}.
Due to the progressive miniaturization and wide-spread availability of camera technology as well as the non-invasive nature of camera-based approaches, most commercially available eye trackers nowadays are video-based (see e.g. \cite{pupil_core,tobii_pro_glasses_2, tobii_pro_glasses_3,ergoneers_dikablis_3,smart_eye_pro,tobii_pro_fusion,eye_link_1000_plus,tobii_pro_spectrum}).
Typically, infrared (IR) cameras and active illumination by IR LEDs are employed.

Eye trackers come in two variants: remote and head-mounted systems.

In remote systems, cameras record the test subject's head and eyes from a fixed location in space, e.g. being attached to a computer screen. While remote eye-tracking systems can achieve excellent accuracy and sampling rates, they restrict movement of the subject to a small region in front of the device \cite{niehorster2018expect}. 
This is a severe limitation, constraining the application space of remote eye-tracking systems.

Head-mounted systems feature near-eye cameras mounted on the head of the test subject, thus recording the eye regions from close-up.
Given a portable recording unit, this setup enables the test subject to move freely in space without affecting the cameras' view of the eyes.
In most setups, additional front-facing cameras allow for correlating the obtained gaze data with environmental cues and stimuli. 
As such, head-mounted eye-tracking systems promise to be unobtrusive and to have the ability to provide meaningful eye-tracking data without constraints on the recording environment and/or the movements of the subject within. 
To date, however, this potential could only be partially realized. 

While the miniaturization of cameras and related technology over the last years has advanced significantly, current head-mounted eye trackers still require rather bulky hardware designs.
These not only tend to obstruct the field of view but also result in an ``odd'' appearance of their wearers. 
The latter has been shown to distort the natural behavior both of the wearers themselves as well as of the people in their surroundings \cite{RisKin11,CanHam18}. 
This behavioral distortion poses a problem for any research seeking to shed light on human cognition and behavior in a natural setting.

In environments that are mostly infrared dark, current generations of head-mounted eye trackers can indeed cope with restricted, low-acceleration movement and work as advertised. 
However, they often suffer from reduced accuracy and missing data in more naturalistic settings. 

Head-mounted eye trackers usually rely on the detection of image features like the pupil contour and/or glints, i.e. reflections actively generated by infrared LEDs.
By causing shadows and environmental reflections, additional infrared light sources, such as the sun, tend to perturb these features and/or altogether hinder their detection within recorded eye images. 
As a consequence, so far eye-tracking studies were mostly limited to indoor environments with sufficiently controlled lighting conditions.

Movement of the eye-tracking headset relative to the subject's head can over time drastically decrease its estimation accuracy \cite{niehorster2020impact}. 
For proper operation, all current commercial eye trackers, including remote systems, require a calibration before each use to determine subject-specific and/or setup-specific parameters.
Not only does this entail a considerable demand on setup complexity and time, but also requires the expertise of test subject and study operator. 
With movement of the headset, the quality of a calibration can deteriorate, effectively necessitating calibration reruns.
Therefore, eye tracking can typically not be done in applications featuring high-acceleration movements, as they encourage headset slippage.

In summary, the perceived social distortion, the lack of robustness to e.g. lighting conditions, as well as the need for a time-intensive setup and repeated calibration, as of today still hamper the widespread use of head-mounted eye trackers in real-world scenarios.

Pupil Invisible glasses, a novel type of head-mounted eye tracker created by Pupil Labs, was designed and engineered with the above limitations in mind (see Fig.~\ref{fig:component_overview}). 
Leveraging recent advances in the field of Deep Learning, it aims at expanding the application space of head-mounted eye tracking. 
With its form factor being practically indistinguishable from a normal pair of glasses, it significantly reduces social distortion. 
At the same time, it claims to offer calibration-free gaze estimation also in truly uncontrolled environments. 

In this work, we present an extensive evaluation of Pupil Invisible glasses and its gaze-estimation pipeline. More specifically, our contributions are the following:

\begin{enumerate}
	\item We provide a high-level overview of the gaze-estimation pipeline employed by Pupil Invisible.
	\item We present a novel geometric framework for gauging gaze-estimation accuracy, which is able to disentangle random errors from biases and resolves gaze-estimation accuracy over the whole field of view.
	\item We evaluate Pupil Invisible glasses on a large in-house data set and present metrics characterizing its expected accuracy in a way that translates directly to real-world usage. In particular, we show that Pupil Invisible glasses provide robust gaze estimation, both in indoor and outdoor lighting conditions, under slippage of the headset, and over the whole range of appearance variations of its wearers.
\end{enumerate}

\section{Related Work}

Our results are related to previous work with regard to (i) techniques for video-based head-mounted gaze estimation and their commercial implementations, (ii) learning-based approaches to the eye-tracking challenge, also beyond the head-mounted setting, and (iii) evaluation schemes for gauging eye-tracker accuracy. 

\subsubsection{Head-Mounted Gaze Estimation}

Over the years, a large variety of approaches for video-based head-mounted gaze estimation has been proposed. Traditionally, these can be distinguished into regression-based and model-based techniques \cite{HanQia10}.

Both variants build on the detection of 2D features, such as pupil centers, pupil contours, or glints in eye images provided by near-eye cameras. In particular, a considerable number of pupil detection algorithms continues to be developed \cite{swirski2012robust, fuhl2015excuse, fuhl2016else, javadi2015set, li2005starburst, santini2017pure}.

Regression-based approaches directly map detected 2D features, typically pupil centers and/or glints, to gaze estimates in a suitable coordinate system defined by a screen or front-facing camera. A common choice for these phenomenological mapping functions is low-order polynomials \cite{CerVil08, HasPey19}. 

Model-based approaches provide gaze predictions based on a 3D eye model, which is fit to the 2D features detected in the eye images, most often pupil contours and glints \cite{SwiDod13,tsukada2012automatic, li2018geometry, li2016two, lai2014hybrid, DieKas18, DieKas19}. 

For proper operation, both types of approaches necessitate a preparatory calibration step prior to each use for determining subject-specific and/or setup-specific parameters tuning the applied gaze-mapping pipeline \cite{long2007high, li2006open}.

Head-mounted eye trackers are commercially available in different hardware configurations from several vendors \cite{pupil_core, tobii_pro_glasses_2, tobii_pro_glasses_3, ergoneers_dikablis_3}. In addition, various prototypes have been proposed by the academic community \cite{san2010evaluation, kim2014development}.
More recently, a number of head-mounted eye trackers integrated into VR/AR headsets have become commercially available \cite{varjo, pupil_vr, htc_vive_pro_eye, fove}.

For most commercial devices, the employed eye-tracking algorithms are proprietary and details have not been disclosed. Pupil Core, an open-source eye tracker sold by Pupil Labs, offers both regression- and model-based modes \cite{pupil_core}. The recently discontinued Tobii Pro Glasses 2 were advertised as making use of a 3D eye model \cite{tobii_pro_glasses_2}.

Here, we present results obtained with Pupil Invisible glasses, a head-mounted eye tracker which circumvents the explicit detection of 2D features and instead performs gaze estimation by directly regressing gaze estimates by means of a convolutional neural network, without necessitating a calibration prior to its use. 

\subsubsection{Learning-Based Gaze Estimation}

Data-driven machine learning (ML) algorithms have superseded classical approaches with respect to a range of computer vision tasks. 
Recently, they have also gained currency within the field of eye tracking. 
In particular, they hold the promise of increasing the robustness of gaze estimation in real-world scenarios. 

One avenue of bringing to bear ML methods in head-mounted gaze estimation, is by substituting classic 2D feature detection algorithms with learning-based counterparts. Indeed,  ML approaches have been successfully employed for pupil center detection \cite{fuhl2016pupilnet, fuhl2017pupilnet} as well as eye segmentation \cite{chaudhary2019ritnet, garbin2019openeds}. 

Another line of work aims at recasting gaze estimation from the bottom up in terms of ML algorithms. Previous studies, however, have almost exclusively focused on remote gaze estimation, starting with earlier works based on traditional ML algorithms \cite{lu2014adaptive, sugano2014learning}, up to more recent contributions building on advances in the field of Deep Learning \cite{zhang2017mpiigaze, zhang2017s, yu2020unsupervised, krafka2016eye}. 

For the head-mounted scenario, a person-specific multi-view approach was formulated  \cite{tonsen2017invisibleeye}. The authors show that by integrating information from several low-resolution near-eye cameras, neural networks can still be successfully employed to regress accurate gaze estimates. 

Exhibiting a considerably different camera setup, Pupil Invisible glasses also utilizes an end-to-end learning-based approach to gaze estimation. The main focus of our work is to provide a thorough validation of its gaze-estimation performance, both using standard and novel quantitative metrics. In particular, our data explicitly attests to the robustness of the Pupil Invisible gaze-estimation pipeline to subject and environmental factors. 

\subsubsection{Evaluation Schemes and Accuracy Metrics}

Suitability of an eye-tracking system for a given research question depends on a range of factors, important ones being required setup time and ease of operation, data-availability in challenging environments, and accuracy of the eye tracker.

Accuracy evaluations provided by manufacturers usually report results obtained in tightly constraint settings \cite{KasPat14, tobii_report}. 
In particular, controlled indoor lighting conditions, limited movement of the subject, and a freshly calibrated system are often assumed. 
These evaluations only partially reflect the real-world performance of the respective eye-tracking system, since in-the-field accuracy is affected by a multitude of factors, such as the recording environment (e.g. outdoor vs. indoor lighting), the recording duration (e.g. due to headset slippage), and subject-specific attributes (e.g. face geometry and eye makeup).

The academic community has provided additional evaluations for a range of commercial eye trackers. While \cite{macinnes2018wearable} provides a comparison in a fairly restricted setting, \cite{niehorster2020impact} is evaluating the impact of slippage of eye-tracking headsets on their performance. Furthermore, \cite{ehinger2019new} is comparing the gaze-estimation accuracy of two eye trackers as well as the quality of derived metrics such as fixation duration.

In this work, we are extending on previous evaluation schemes in several ways. 
Earlier evaluations were performed on fairly small subject pools, ranging from 3 to 20 subjects \cite{KasPat14, tobii_report,macinnes2018wearable,ehinger2019new,niehorster2020impact}.
Here, we present accuracy results for N=367 unique validation subjects. Furthermore, during data collection we (i) have made efforts to mimic the real-world usage of Pupil Invisible glasses, (ii) have realized natural slippage configurations, (iii) have obtained relevant metadata (age, gender appearance, etc.) for each validation subject, and (iv) did not pre-filter any data on the subject or sample level. This not only permits us to present sound population-level statistics representative of in-the-field usage of Pupil Invisible glasses, but also to probe the effects of various subject-specific factors on gaze-estimation accuracy. 
In particular, we propose a novel framework for analyzing gaze estimation accuracy which extends on the common metric of mean angular error. In doing so, we resolve gaze-estimation accuracy over the field of view and shed light on biases and directional sensitivities of the gaze estimates provided by Pupil Invisible glasses. 

\section{Pupil Invisible Glasses}

We start with a high-level overview of the hardware and gaze-estimation pipeline employed by Pupil Invisible glasses.

\subsection{Hardware Design}

Pupil Invisible glasses consist of a 3D-printed frame resembling a regular pair of glasses (see Fig.~\ref{fig:component_overview}). Two IR near-eye cameras are fully embedded temporally into the frame, one on each side. An IR LED in close proximity provides sufficient illumination of the respective eye region, also in IR-dark environments. 
A scene camera with a field of view of about $90^\circ\times 90^\circ$ can be attached to the left temple via a magnetic connector. 

The frame is slightly flexible, which allows it to be worn comfortably by subjects with a wide range of head sizes. 
For subjects with heads too small for a snug fit, e.g. children, an additional head strap can be used to fixate the glasses.

Due to the fully embedded design, Pupil Invisible glasses do not immediately stand out as an eye tracker, but rather appear like regular prescription glasses. 
Thus, they are expected to create little social obstruction for the wearer and their surrounding, allowing them to behave naturally. 
Furthermore, since no camera adjustment is needed, setup time is considerably reduced. 

Pupil Invisible glasses are not equipped for on-device computation. 
Instead, in order to perform gaze estimation, they need to be cable-connected to a smartphone, referred to as the Pupil Invisible Companion device. 
The latter is handling all computation and storage of the recorded data.

Pupil Invisible glasses offer several other sensors and features. 
As those are independent of the gaze estimation pipeline, however, they will not be considered here. 
A complete description of Pupil Invisible glasses can be found at \url{www.pupil-labs.com}.

\subsection{Gaze-Estimation Pipeline}

\begin{figure*}[t]
	\centering
	\includegraphics[width=\textwidth]{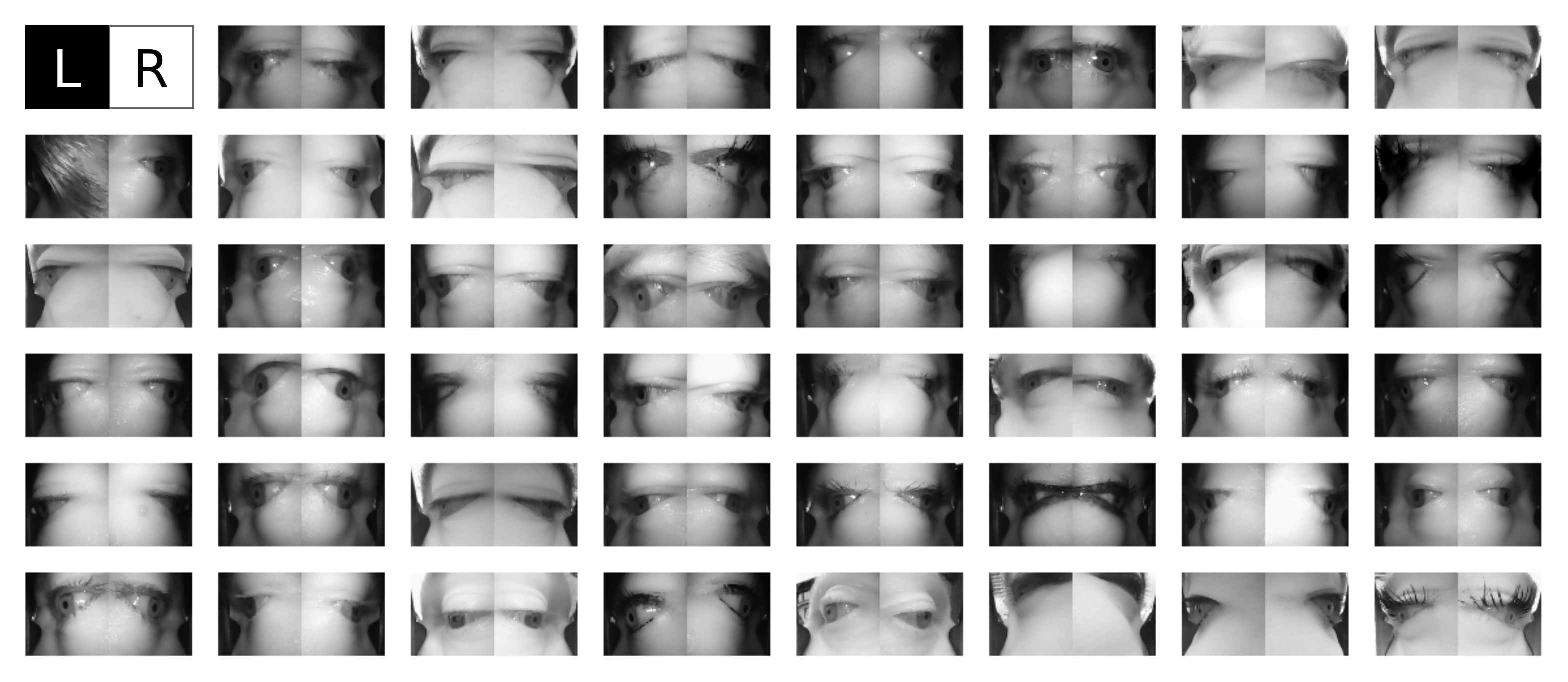}
	\caption{\label{fig:sample_images}
		Sample pairs of eye images recorded with Pupil Invisible glasses. 
		All samples shown consist of two concurrent images, provided by the left and right eye camera, respectively, and are part of an in-house data set recorded by Pupil Labs. 
		During recording, subjects were asked to focus their gaze on a fixed marker from various angles and distances. 
		The 3D gaze point was then determined by automatically detecting and analyzing the marker image in the scene video. 
		Here, samples were chosen for a fixed 3D ground-truth label ($p_{\rm gt}=(0\,{\rm cm},0\,{\rm cm},300\,{\rm cm})$). 
		Note the variation of face geometries and lighting conditions.}
\end{figure*}

Gaze estimation by Pupil Invisible glasses is performed on the basis of individual pairs of concurrent eye images $(e_\ell, e_r)$, taken by the left and right eye camera, respectively (see Fig.~\ref{fig:sample_images}).
The quantity being predicted is referred to as gaze direction, a unit vector pointing from the origin of a suitable coordinate system to the 3D gaze point. 
More specifically, the final device-specific gaze-direction estimate $d_{\rm dev}$ is expressed in the coordinate system set up by the scene camera and is reported as a 2D gaze point $p_{\rm dev}$ in pixel space. 
Note, assuming a camera projection model and given the corresponding intrinsics of the scene camera, gaze direction and 2D gaze point are merely different representations of the same geometric quantity.

The gaze estimation pipeline $\mathcal{G}$ employed by Pupil Invisible glasses consists of two steps: (i) estimating the gaze direction  $d_{\rm ideal}$ in a \emph{device-independent} coordinate system, and (ii) mapping the gaze-direction estimate $d_{\rm ideal}$ to \emph{device-specific} scene camera pixel space, resulting in a final 2D estimated gaze point $p_{\rm dev}$. We will deal with these two steps in turn. 

i) Due to tolerances in the manufacturing process of Pupil Invisible glasses, both the relative extrinsics of the eye cameras and scene camera (i.e. their spatial relationships), as well as the intrinsics of the scene camera module vary slightly from hardware instance to hardware instance. 
For each hardware instance, the respective quantities are measured during production and thus can be assumed to be known. 
In order to compensate for these production variances, in a first step, gaze direction $d_{\rm ideal}$ is estimated in a device-independent coordinate system. 
The latter can be thought of as being defined by an idealized scene camera.

Predictions are obtained by means of a convolutional neural network that was trained on a large and diverse in-house data set recorded by Pupil Labs, consisting of pairs of concurrent eye images and corresponding 3D ground-truth gaze labels $p_{gt}$.
The architecture of the employed neural network was specifically designed to be efficient enough to be executed on a mobile device in real-time.
By utilizing the built-in GPU of Pupil Invisible Companion phones, forward-passes through the network can be calculated at a rate of 55 Hz.

ii) In a second step, device-specific extrinsics and intrinsics are used to transform the gaze direction obtained in (i) to a gaze direction $d_{\rm dev}$, which is specific to the actual scene camera module mounted on the employed Pupil Invisible glasses instance. Projection into pixel space (using the measured intrinsics) results in the final estimated 2D gaze point, $\mathcal{G}(e_\ell, e_r)=p_{\rm dev}$. 

Note, no calibration step is necessary for recording gaze data with Pupil Invisible glasses. 

\section{Accuracy Metrics}
In this section, we present the evaluation scheme employed for assessing the gaze-estimation accuracy of Pupil Invisible glasses.
We begin with a description of the validation data set and its recording protocol.
We then introduce the metrics used to assess subject- and population-level statistics pertinent to gaze-estimation accuracy. 

\subsection{Validation Data}
\label{sec:validation_data}
To serve as validation subjects, we randomly chose N=367 unique subjects from our in-house data set. 
In order to obtain meaningful and challenging validation data, we devised a recording protocol and dynamic choreography reflective of the real-world usage of Pupil Invisible glasses. 
In particular, over the course of several recordings, each validation subject fixated a marker from a large variety of angles and depths. 
From these recordings, we distilled triples $(e_\ell, e_r, p_{\rm gt})$, corresponding to concurrent left and right eye images and ground-truth 3D gaze points, also referred to as gaze labels, in device-specific scene camera coordinates, respectively (cf. sample images in Fig.~\ref{fig:sample_images} for a fixed choice of $p_{\rm gt}$).
The device-specific ground-truth gaze direction $d_{\rm gt}$ can be obtained from $p_{\rm gt}$ by vector normalization. 
Note, no subjects were discarded and no filtering of gaze samples was applied post recording, i.e. 100\,\% of the recorded gaze samples, including blinks and extreme gaze directions, were taken into account when obtaining the results presented below.  

\begin{figure}[t]
	\centering
	\includegraphics[width=\columnwidth]{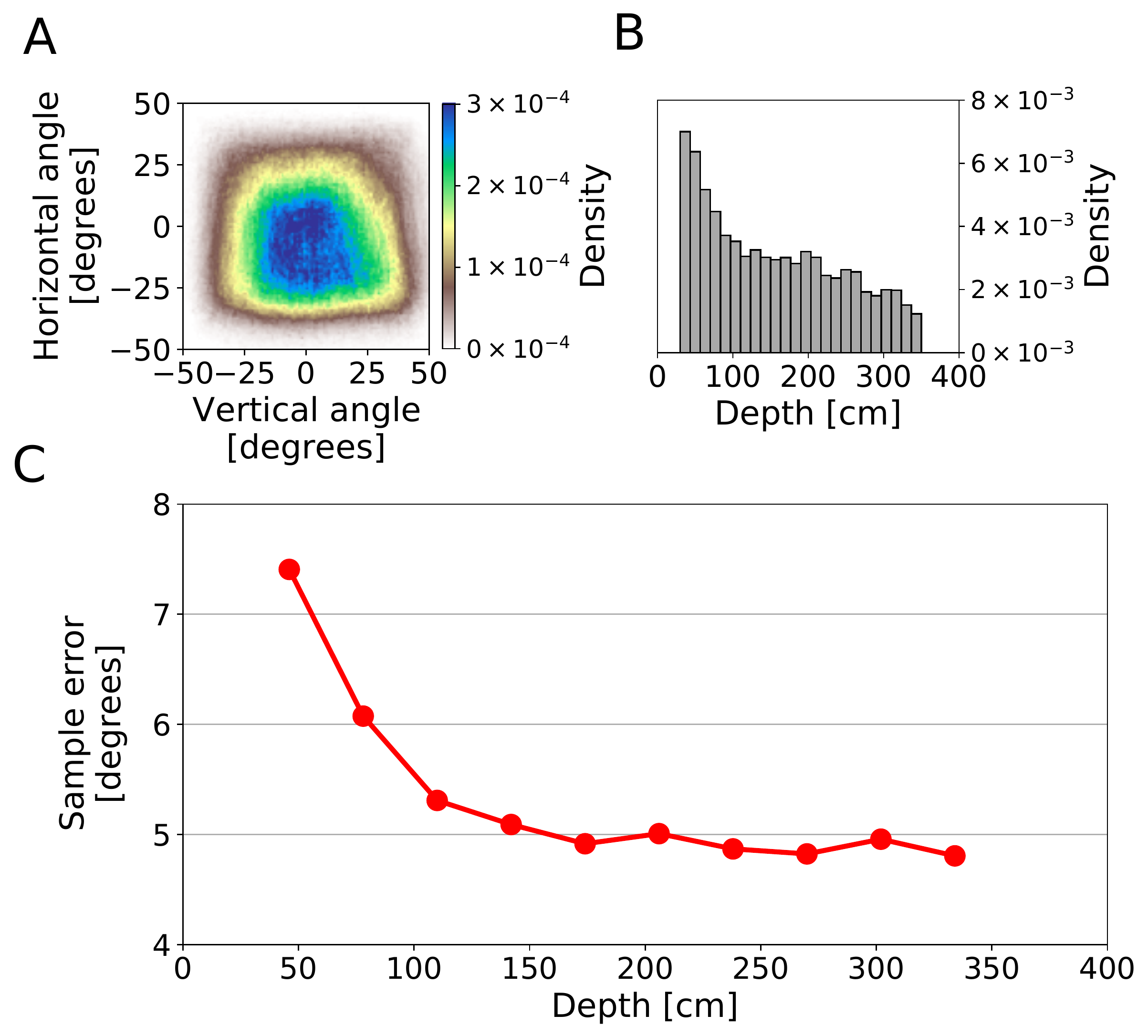}
	\caption{\label{fig:depth_distribution} Density of ground-truth gaze targets in the validation data set, shown in (A) as a function of horizontal and vertical angle, and in (B) as a function of depth. Covering a field of view of $\approx90^\circ\times90^\circ$, ground-truth gaze targets exhibit a slight center bias. Due to the employed recording protocol, gaze targets at distances below $\approx 100$ cm are slightly over-represented. (C) Mean sample error of gaze estimates averaged over all validation subjects as a function of depth. Being constant beyond $\approx150$ cm, mean sample error increases for shorter gaze depths ($\approx40$ \% at $50$ cm).}
\end{figure}

In Fig.~\ref{fig:depth_distribution}A and B, we show plots of the resulting label density aggregated over all subjects in the validation data set. 
Gaze samples cover a range as large as about $\pm45^\circ$ horizontally and vertically, i.e. the whole field of view of the attached scene camera. 
Note that due to the recording protocol, the gaze sample density exhibits a slight center bias. 
Gaze samples were recorded in a depth range of approximately 30 cm to 350 cm, with gaze samples at short distances being slightly over-represented. 
Note, when fixing gaze direction, changes in eye vergence beyond 350 cm are negligible.
Our accuracy results thus also reflect accuracy at depths larger than the maximum depth recorded. 

For each subject, gaze samples were recorded in an outdoor and indoor environment, with the indoor environment being IR-dark and the outdoor environment ranging from cloudy to sunny (cf. sample images in Fig.~\ref{fig:sample_images}). 
In order to increase the range of naturally occurring slippage configurations, each subject was asked to manually shift and randomly readjust the position of Pupil Invisible glasses between recordings.
Metadata was collected for each subject as to their age, gender appearance, interpupillary distance (IPD), usage of contact lenses during the recording, and the presence of eye makeup.

As expounded in the last section, gaze estimation in Pupil Invisible glasses comprises a device-specific projection step.
In order to obtain results representative of in-the-field usage of Pupil Invisible glasses, test subjects were grouped according to the specific Pupil Invisible hardware instance used during recording. 
For each group, the Pupil Invisible gaze-estimation pipeline was trained on the remaining subjects in our in-house training data set. 
In particular, the training data never contained samples recorded with the same pair of Pupil Invisible glasses as used during evaluation.
For each pair of concurrent eye images in the validation data set, we then obtained the corresponding predicted gaze point $p_{\rm dev}$ and equivalent gaze direction $d_{\rm dev}$. 

\subsection{Subject-Level Statistics}

Given a gaze sample $(e_\ell, e_r, p_{\rm gt})$ with corresponding ground-truth gaze direction $d_{\rm gt}$ and predicted gaze direction $d_{\rm dev}$, we refer to the angle between $d_{\rm gt}$ and $d_{\rm dev}$ as the \emph{sample error}. 
In order to facilitate the analysis of environmental and subject-specific factors on gaze-estimation accuracy, we aggregate sample errors for a given subject in a given condition into a single number, referred to as \emph{subject error}. 
Note, sample error is not uniform in space. 
In particular, we found that mean sample error (averaged over the whole validation data set) tends to be larger at smaller depths (see Fig.~\ref{fig:depth_distribution}C). 
Since gaze samples are not distributed uniformly over depth (see Fig.~\ref{fig:depth_distribution}B), we opted for a binning strategy.
More specifically, we define subject error as the mean of mean gaze error as calculated in ten depth bins, uniformly distributed between 30 cm and 350 cm. 
Note, no other normalization of gaze sample distributions was performed. 
Reported subject errors thus exhibit a slight dependence on the actual set of gaze samples recorded for a given subject in a given condition.

\subsection{Population-Level Statistics}
Aggregated statistics like the subject error defined in the last section are appropriate means for probing the effect of environmental and subject-specific factors on gaze-estimation accuracy (see results in section \ref{sec:results_subject_error}). 
A drawback of the utilized subject-level metric is, however, that it does not resolve gaze-estimation error over the field of view. 
Neither does it provide insights as to the type of error made by the gaze-estimation pipeline, i.e. whether the error takes the form of a bias (shifted mean gaze prediction) or rather of a spread of the gaze estimates around a local ground-truth gaze direction. 
In order to shed light on these questions, in this section we develop a framework for characterizing gaze-estimation accuracy on a population level, which allows for this desired differentiation. \begin{figure*}
	\centering
	\includegraphics[width=0.95\textwidth]{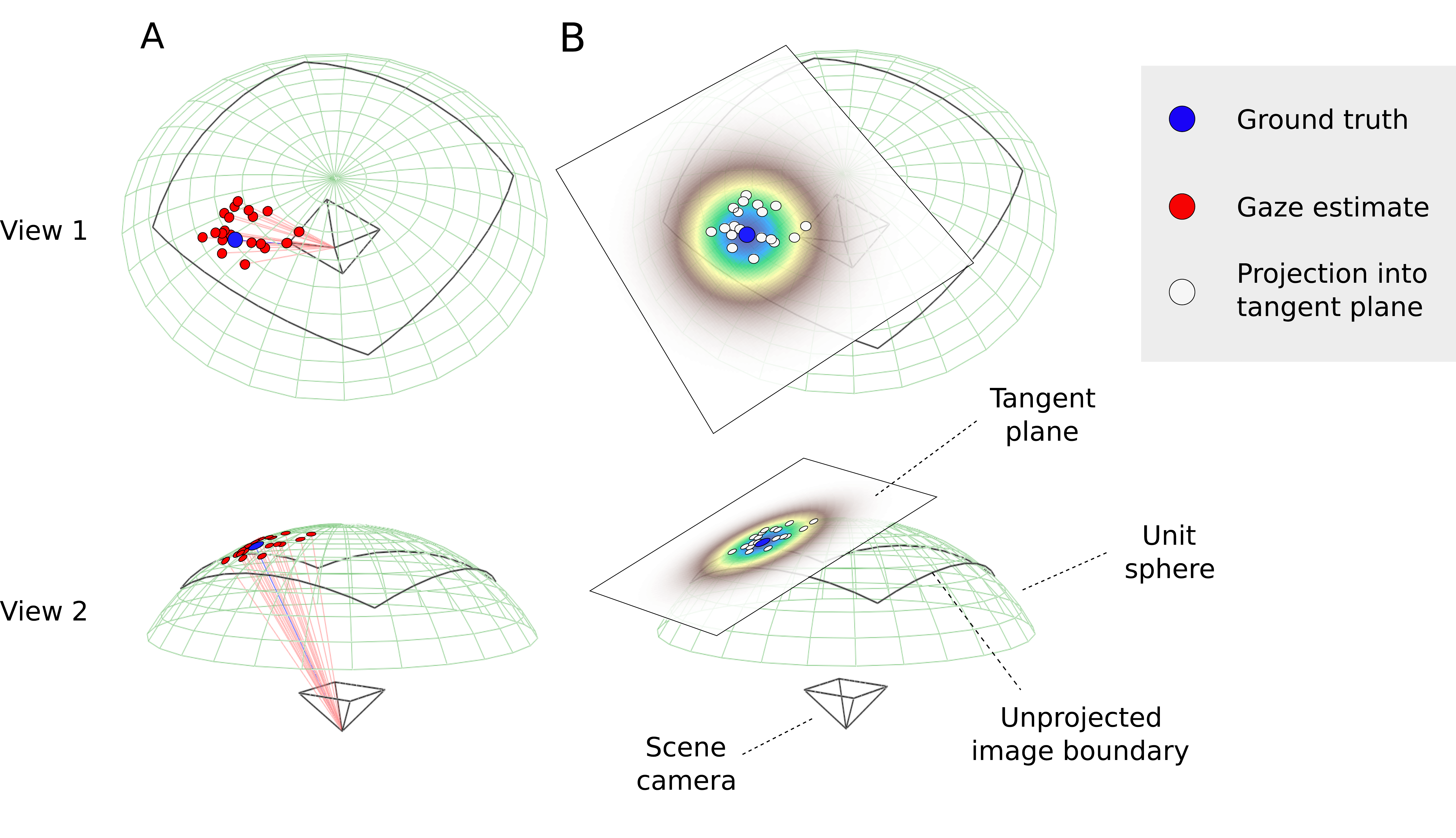}
	\caption{\label{fig:sketch_tangent_plante}(A) When expressing the ground-truth gaze direction $d_{\rm gt}$ in spherical coordinates, it can be visualized as a point on the sphere (see blue disk in upper and lower panel; both panels show the same sphere section, albeit from different viewpoints). Application of the Pupil Invisible gaze-estimation pipeline to corresponding pairs of concurrent left and right eye images results in gaze estimates $d_{\rm dev}$, which are distributed around the ground-truth value (red disks in upper and lower panel). (B) Orthogonal projection into the local tangent plane effectively flattens out the curvature of the sphere. Gaze estimates can thus be viewed as points in a 2D plane (white disks). As such, they give rise to a 2D distribution of points, which can be accounted for by a 2D Gaussian (density shown as heatmap).} 
\end{figure*}
To this end, assume a set of pairs of eye images $(e_\ell, e_r)$, recorded by a number of subjects in various environments, which all correspond to the same ground-truth gaze direction $d_{\rm gt}$. 
Application of our gaze pipeline $\mathcal{G}$  results in a corresponding set of predicted gaze directions $d_{\rm dev}$, i.e. $\mathcal{G}(e_\ell, e_r)=d_{\rm dev}$. Since both $d_{\rm gt}$ and $d_{\rm dev}$ are normalized they can be jointly visualized as points on the unit sphere (see Fig.~\ref{fig:sketch_tangent_plante}A). 
Note, perfect gaze estimation would imply $\mathcal{G}(e_\ell, e_r)\equiv d_{\rm gt}$. 
In practice, however, the predictions $d_{\rm dev}$ will be distributed on the sphere around $d_{gt}$ (we will present an example based on our validation data set in Fig.~\ref{fig:fit_example}).

To transform the resulting distribution on the sphere into a 2D distribution, we map predicted gaze directions to coordinates defined within the plane tangent to the sphere in $d_{\rm gt}$ (see Fig.~\ref{fig:sketch_tangent_plante}B). 
We denote this tangent plane by $t_{\rm gt}$. 
More specifically, we introduce a coordinate system in $t_{\rm gt}$ by choosing normalized vectors $b_1, b_2$ in $\mathbb{R}^3$, such that $(d_{\rm gt},b_1,b_2)$ constitutes an orthonormal basis of $\mathbb{R}^3$. 
We then consider the mapping $\mathbf{T}_{\rm gt}$ with 
\begin{equation}
\mathbf{T}_{\rm gt}(d_{\rm dev}; b_1, b_2) = \left(\begin{array}{cc} b_{11} & b_{12} \\ b_{21} & b_{22} \end{array}\right)d_{\rm dev}.
\end{equation}
Note, $\mathbf{T}_{\rm gt}$ is unique only up to an arbitrary rotation.
In geometric terms, one can think of $\mathbf{T}_{\rm gt}$ as flattening the sphere around $d_{\rm gt}$. 
In particular, for points within a couple of degrees of $d_{\rm gt}$, this mapping introduces negligible distortion. 

Statistical properties of gaze-direction estimates can thus be studied by an analysis of the distribution of mapped points instead. 
We propose a fit with a 2D Gaussian (see Fig.~\ref{fig:sketch_tangent_plante}B). 
For a refresher on 2D Gaussian distributions and their geometric interpretation, we refer the reader to the Appendix. 
In particular, a 2D Gaussian is characterized by a mean value $\mu$ and a covariance matrix $\mathbf{\Sigma}$.
From a geometrical point of view, a 2D Gaussian can be visualized as a shifted ellipse, where the shift of the ellipse center is given by $\mu$ and its orientation and minor and major axis are encoded by $\mathbf{\Sigma}$.
For a fixed $d_{\rm gt}$, characterizing the resulting distribution of gaze estimates by means of a shifted ellipse resolves the local gaze-estimation error in terms of a bias (represented by the shift $\mu$) and/or a directionally skewed random error characterized by the orientation and extent of its minor and major axis (represented by the covariance matrix $\mathbf{\Sigma}$).
Performing the proposed analysis over a grid of ground-truth directions allows for diagnosing gaze-estimation accuracy resolved over the whole field of view. 
We refer to this population-level statistics as \emph{directional statistics}.

\section{Results}
Utilizing the accuracy metrics introduced in the last section, this section presents quantitative results assessing the gaze-estimation accuracy of Pupil Invisible glasses. 
By providing subject-level statistics in terms of subject error, we show that gaze-estimation accuracy is robust to environmental and subject-specific factors. 
Obtaining population-level results in terms of directional statistics, we characterize gaze-estimation accuracy resolved over the whole field of view. 
\begin{figure}[h!]
	\centering
	\includegraphics[width=0.84\columnwidth]{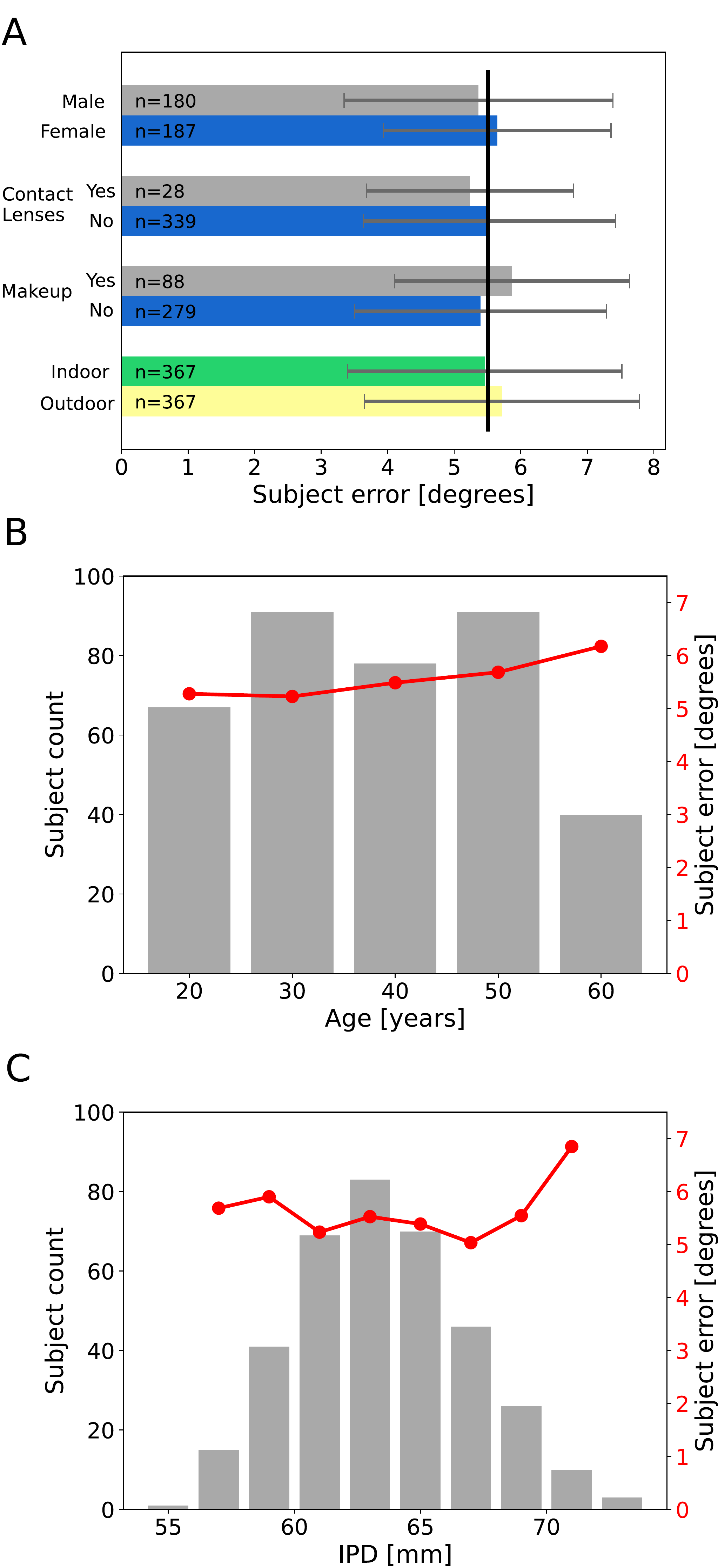}
	\caption{\label{fig:subject_error} Statistics of subject error. (A) Comparison of subject errors calculated over various splits of the validation data. 
		Note that corresponding gray and blue bars differ in the underlying ensemble of subjects. 
		The green and yellow bars represent the same subjects, but differ in the environment in which the validation data was recorded.
		Error bars denote standard deviations and comprise effects from inter-subject variability and the subject-specific gaze sample distribution. The number of subjects making up each group is indicated. In all cases, we find that mean values and standard deviations are similar. 
		(B) Mean subject error as a function of subject age (red line) and age histogram (gray bars). 
		(C) Mean subject error as function of IPD (red line) and IPD histogram (gray bars). 
		As our data shows, mean subject error does neither depend on age nor on IPD.}
\end{figure}
\subsection{Subject Error}
\label{sec:results_subject_error}
Based on the metadata available for each subject, we split the subject pool along various dimensions: gender appearance, presence of contact lenses, and presence of eye makeup. 
For each split, we determined the mean subject error and its standard deviation. 
As can be seen from the data in Fig.~\ref{fig:subject_error}A (gray and blue bars, number of subjects per split as indicated), mean subject error between splits does differ only marginally and is $\approx 5.5^\circ$ in all cases. 
As to the standard deviation, we also find it to be consistent between splits ($\approx2.0^\circ$). 
Note, variations in subject error are determined both by intrinsic inter-subject differences as well as differences introduced due to the non-uniform sampling of gaze targets during recording (see section \ref{sec:validation_data}). 
To answer whether lighting conditions have an effect on subject error, we calculated subject error for each subject once for the recordings obtained indoors and once for the recordings performed outdoors (see Fig.~\ref{fig:subject_error}A, green and yellow bar). 
Again we find no noteworthy difference in mean values and standard deviations, respectively, of the resulting subject errors ($\mbox{mean}\approx 5.5^\circ$, $\mbox{std}\approx 2.0^\circ$).

Lastly, we binned subjects according to IPD and age (see Fig.~\ref{fig:subject_error}B and C). 
For each respective bin, we calculated the mean subject error and subject count. 
Ages between 18 and 64 years are almost uniformly distributed in our validation data set.
The distribution of measured IPDs is peaked around a mean value of $63.2$ cm and is comparable to similar results reported in the literature \cite{Dod04}.
Neither for age nor for IPD does the data show a strong variation between respective groups (see red lines in Fig.~\ref{fig:subject_error}B and C).

In summary, the results presented in this section show that gaze-estimation accuracy as offered by Pupil Invisible glasses is independent of lighting conditions and subject-specific factors (age, gender appearance, IPD, makeup, contact lenses). 
Since recordings for each subject also include variations of the pose of Pupil Invisible glasses relative to the head, these results also give a clear indication of what gaze-estimation accuracy is expected in slippage-prone real-world scenarios. 
Note, no data was discarded in our evaluation, i.e. statistics reflect accuracy as averaged over the whole field of view and 100\,\% of the available gaze samples. 

\subsection{Directional Statistics}
\begin{figure}[t]
	\centering
	\includegraphics[width=\columnwidth]{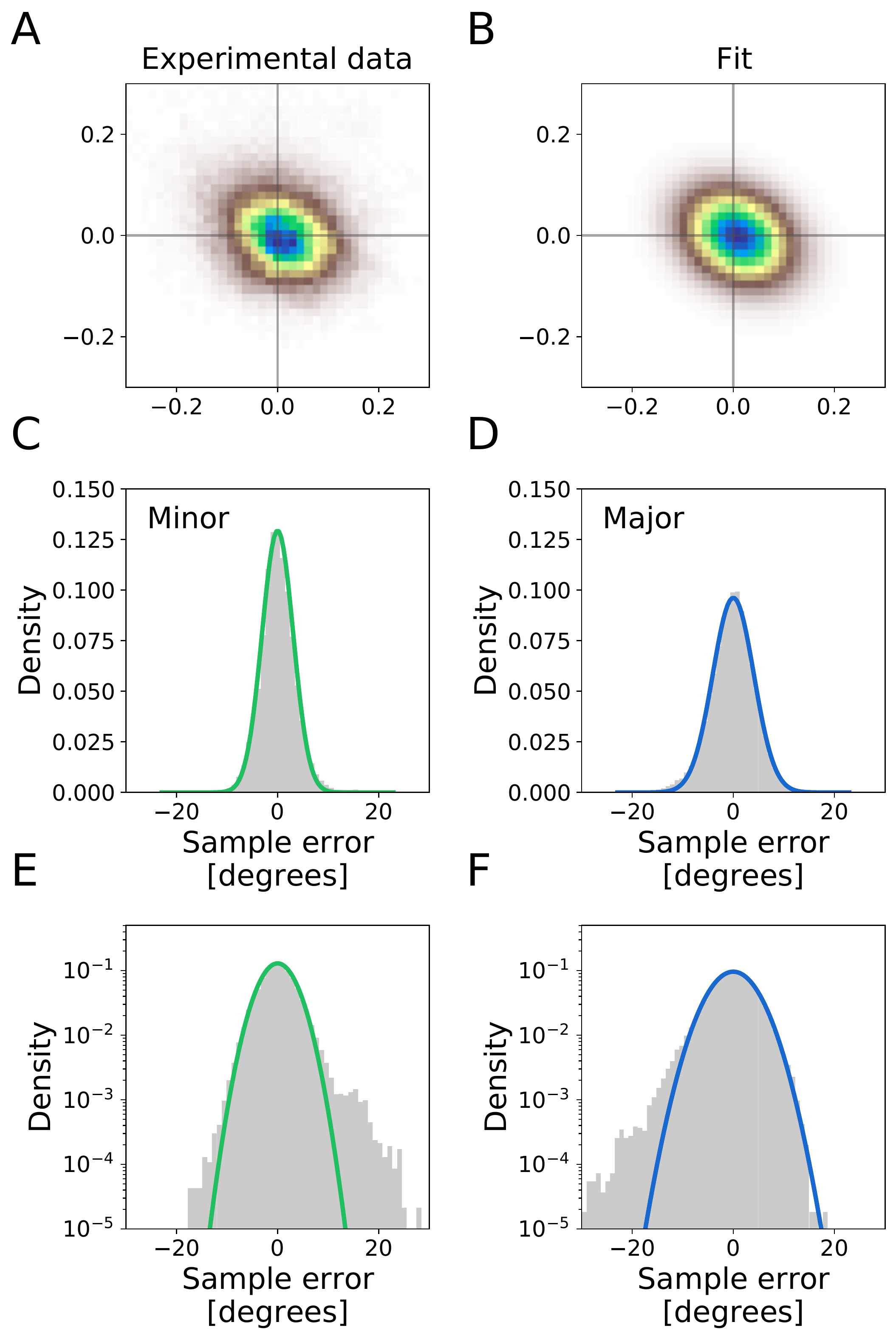}
	\caption{\label{fig:fit_example} Distribution of gaze estimates in tangent space. (A) Normalized histogram of gaze estimates after projection into the local tangent plane $t_{\rm gt}$. (B) Resulting fit of the distribution in A with a 2D Gaussian. (C) \& (D) Normalized histograms of gaze estimates along the minor and major axis, respectively. Predictions from the fit with a 2D Gaussian are shown as green and blue lines, respectively. (E) \& (F) Same data as shown in C and D, albeit on a logarithmic scale. Note that due to outliers, the distribution of gaze estimates deviates slightly from the exponential falloff of the fitted distribution. In order to reduce the influence of these outliers on the fitting result, we opted for a robust fitting procedure.}
\end{figure}

We start by showing an example distribution of gaze estimates in the local tangent space and the corresponding fit with a 2D Gaussian distribution (see Fig. \ref{fig:fit_example}). 

To this end,  consider $d_{\rm gt}=(0.276, -0.276,  0.921)$, a gaze direction in the upper right quadrant of the field of view. 

Note, in practice no gaze sample will \emph{exactly} correspond to $d_{\rm gt}$. 
Thus, gaze samples in close proximity to $d_{\rm gt}$ need to be considered and the residual offset needs to be corrected for. 
More specifically, we started by selecting all samples $(d_{\rm gt}^i, d_{\rm dev}^i)$ from the validation data set for which $d_{\rm gt}^i$ deviates from $d_{\rm gt}$ by less than five degrees. 
To correct for the angular mismatch, we determined the rotation $\mathbf{R}^i$ around an axis orthogonal to $d_{\rm gt}$ and $d_{\rm gt}^i$, which rotates $d_{\rm gt}^i$ onto $d_{\rm gt}$, i.e. for which $\mathbf{R}^id_{\rm gt}^i=d_{\rm gt}$. 
Application of $\mathbf{R}^i$ then yielded offset-corrected gaze estimates 
\begin{equation}
\tilde{d}_{\rm dev}^i=\mathbf{R}^i d^i_{\rm dev}.
\end{equation}
For ease of notation, we will omit the tilde in the following. 
Offset-corrected gaze estimates $d_{\rm dev}^i$ were then projected into the local tangent space.
Note, $d_{\rm gt}$ maps to the origin by construction.

As can be seen from Fig.~\ref{fig:fit_example}A, gaze estimates are distributed asymmetrically around the ground-truth value, with the center of mass being close to the origin (cf. intersection of gray lines). 
The density of gaze estimates is well fit by a Gaussian distribution (see Fig.~\ref{fig:fit_example}B).
In practice, we opted for a least-squares fit, made robust to outliers by inclusion of a Cauchy loss in the fitting objective.
The quality of the fit also becomes apparent when considering histograms along the minor and major axis of the distribution (see Fig.~\ref{fig:fit_example}C and D; a definition is given in the Appendix). 
As expected, the density along the minor axis is more peaked than along the major axis. 
For completeness, we also show the same histograms and fits on a logarithmic scale (see Fig.~\ref{fig:fit_example}E and F). 
Note, due to the existence of outliers, the decay of the histograms for large angular errors deviates slightly from the exponential falloff of the Gaussian distribution. 
The observed outliers are most likely due to blinks and intermittent failures of validation subjects to fixate the marker. 

We conclude that the proposed analysis via 2D Gaussian distributions in the local tangent plane is feasible and descriptive of the distribution of gaze estimates in our validation data set. 
\begin{figure*}[h!]
	\centering
	\includegraphics[width=\textwidth]{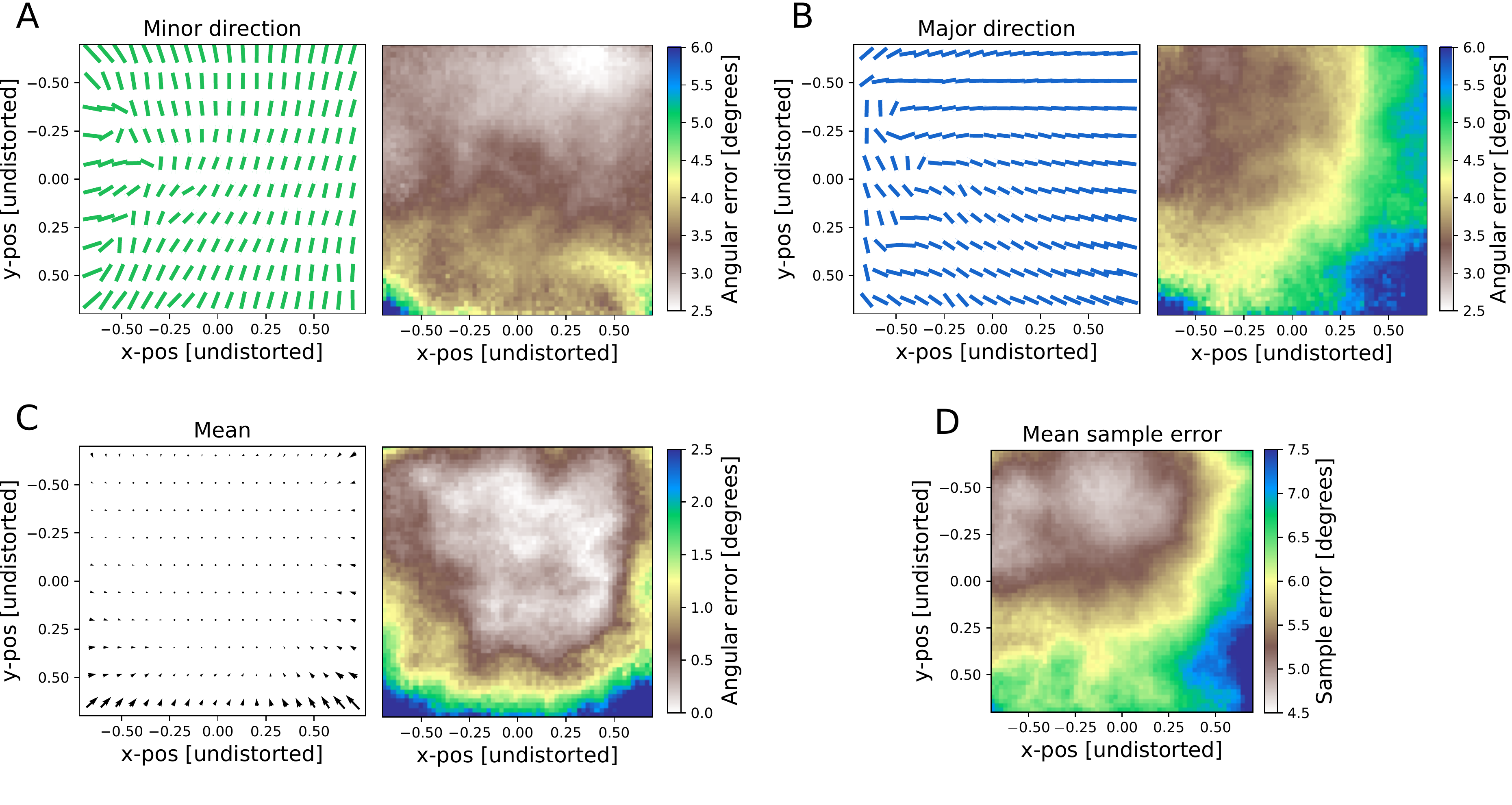}
	\caption{\label{fig:dc_directional_statistics} Directional statistics of gaze estimates in the validation data set. (A) \& (B) Projection of the local minor and major axis, respectively, into normalized image space. Axis directions are shown as green (minor axis) and blue (major axis) lines (left panel). The length of the respective axes are converted to visual angle and are shown as a heatmap (right panel). (C) Projections of the mean vector as calculated in the local tangent space (black arrows; left panel). For clarity, we show the length of the mean vectors after conversion to visual angle as a heatmap (right panel). (D) Mean sample error as calculated in the local tangent space as a function of undistorted image space coordinates. }
\end{figure*}
In order to shed light on Pupil Invisible gaze-estimation accuracy as a function of ground-truth gaze direction, we have repeated the above analysis for a grid of gaze directions, covering the whole field of view of the scene camera. 
Results of this analysis are presented in Fig.~\ref{fig:dc_directional_statistics}.

For each point within the grid, we determined the minor and major axis as described above. 
In order to graphically present our results, we project the respective quantities (ground-truth gaze direction, minor and major axis) into normalized image coordinates using the camera matrix of a typical scene camera.
We refer to this space as undistorted image space. 
More specifically, in Fig.~\ref{fig:dc_directional_statistics}A and B, we indicate the orientation of the axes as a function of undistorted image space coordinates (green lines for minor, blue lines for major axis). 
Standard deviations along the respective axis are shown as a heatmap and referred to as angular error (with lengths in tangent space converted to visual angle). From these plots, we conclude that gaze estimation over most of the field of view exhibits less random error in the vertical than in the horizontal direction, with errors in the center of the field of view being as a low as about $3^\circ$ in the vertical, and about $4^\circ$ in the horizontal direction.
Note, minimal values in the vertical direction (minor direction) are achieved in the upper half of the field of view (down to about $2.5^\circ$), in the horizontal direction (major direction) in the upper left quadrant (down to about $3.5^\circ$). 
Random errors in both cases are slightly higher in the lower third of the field of view. 
Note, by construction the errors reported in panel B are point-wise higher than the ones reported in panel A. 
Since major and minor axis are orthogonal in tangent space, this is also the case after projection to undistorted image space. 

In order to unravel whether gaze estimation by Pupil Invisible glasses is prone to biases, we visualized the extracted mean vectors. 
In Fig.~\ref{fig:dc_directional_statistics}C, we present a quiver plot with arrows pointing from the ground-truth gaze direction to the mean of the fitted 2D Gaussian, i.e. the mean predicted gaze direction. 
The induced angular error due to this shift is shown in units of visual angle as a heatmap. 
As can be seen from the data, over the whole horizontal and large parts of the vertical field
of view, estimation bias is lower than $0.5^\circ$. 
In the lower fifth of the field of view, it increases to about $2.5^\circ$. 
Only considering their direction, arrows are pointing towards the center everywhere, indicative of a slight center bias of the Pupil Invisible glasses gaze-estimation pipeline. 

For the sake of comparison, we also show the mean sample error as a function of undistorted image space coordinates (see Fig.~\ref{fig:dc_directional_statistics}D). 
As in panel B, minimal values are achieved in the upper-left quadrant, where mean sample error is about $5^\circ$. 
In the lower quarter of the field of view, mean sample error increases slightly to about $6.5^\circ$. 
Note, mean sample error is a composite of all five parameters characterizing a 2D Gaussian distribution (see Appendix). 
In particular, it is a poor estimator for the extent of random scatter of samples around the mean value. 
More precisely, it can be shown that it overestimates the effective spread, even in the limit case of a symmetric distribution with vanishing mean vector. 

Our analysis shows that Pupil Invisible glasses deliver unbiased gaze estimates with less than about $4^\circ$ of random spread over most of the field of view, without the need of any calibration. 
Due to head movements as well as deliberate readjustments, validation recordings include movements of Pupil Invisible glasses relative to the head. 
Thus, our results also attest to the robustness of the employed gaze-estimation pipeline to headset slippage. 
The expected spread of gaze estimates is lower in the vertical than the horizontal direction. 
Vertical gaze estimation is slightly more prone to error in the lower fifth of the image. 
This is most likely related to the occlusion of the eye by eye-lashes, which is most pronounced when subjects are looking downward. 
Even for these extreme gaze directions, however, the observed random spread rarely exceeds values of $4.5^\circ$. 
For corresponding gaze directions, a slight center bias can be discerned, which is likewise suggestive of the reduced correlation in eye-appearance and vertical gaze direction. 
Random error in the horizontal direction is minimal in the upper-left quadrant ($\approx 3.5^\circ$). 
Inspection of recorded eye images suggests, pupil visibility for the left eye is best in this quadrant (data not shown).  
Our data thus indicates that the gaze pipeline of Pupil Invisible glasses is sensitive in particular to pupil visibility in the left eye image. 

\section{Discussion}

In this work, we have provided an evaluation of Pupil Invisible glasses, a calibration-free head-mounted eye tracker recently introduced by Pupil Labs. 
Due to its unobtrusive form factor, Pupil Invisible glasses are expected to reduce the risk of social distortion influencing subject behavior during recording sessions.
Given its ease of use and reduced setup time, Pupil Invisible glasses promise to enable novel use cases of head-mounted eye tracking.
We have given a high-level overview of its relevant hardware components and have detailed important facts with regard to the employed gaze-estimation pipeline. 
Adopting a learning-based approach, Pupil Invisible glasses bring to bear a convolutional neural network trained on a large in-house data set for providing gaze estimates based on pairs of concurrent eye images. 
Initial gaze estimates are expressed in an ideal scene camera space and then adjusted in a projection step to the actual hardware instance on the basis of factory-provided hardware-specific calibration data.

We have presented an extensive evaluation of the gaze-estimation capabilities of Pupil Invisible glasses. 
To this end, we have devised an evaluation scheme shedding light on the accuracy of gaze estimates during real-world usage. 
Based on recordings from N=367 subjects and comprising subject- and population-level statistics, we have shown that Pupil Invisible glasses deliver gaze estimation, which is robust to headset slippage, as well as environmental and subject-specific factors. 
In order to resolve gaze-estimation accuracy over the field of view and gain insight into the type of error introduced by the gaze-estimation pipeline, we have developed a novel statistical framework. 
The abundance of available validation data allowed us to analyze the distribution of gaze estimates in local tangent planes. 
By fitting the measured distributions with 2D Gaussians, we were able to quantify the relative contributions of biases and random spreads to gaze-estimation errors. 
Gaze estimation was shown to be less prone to random error in the vertical than in the horizontal direction. 
In particular, we could show that gaze estimates are unbiased over large portions of the field of view, exhibiting a random spread of only about $4^\circ$. 
We believe that the proposed statistical framework holds the potential to serve as a powerful tool for characterizing eye-tracking accuracy also beyond the evaluation shown here.

While Pupil Invisible glasses operate robustly without the need for a calibration step by the user, there is a limit as to what level of gaze-estimation accuracy can be achieved with calibration-free approaches. 
This is mainly due to the fact that not all relevant physiological characteristics of the human eye are accessible by mere image analysis. 
An important example is the person-specific offset between the visual and the optical axis, which cannot be determined from eye images alone. 
Being forced to rely on average values, calibration-free methods by design cannot modulate estimated gaze directions along with all relevant person-specific parameters. 
It thus stands to reason that in order to overcome the limits of calibration-free approaches, calibration schemes for inferring and utilizing such subject-specific parameters are unavoidable. 
As a step in this direction, within the Pupil Invisible Companion app users can manually determine a subject-specific constant gaze offset which is found to further boost gaze-estimation accuracy in many cases.
Also, several groups, both in academia and industry, have undertaken promising efforts to incorporate calibration steps into otherwise black-box learning-based approaches \cite{YuLiu19, ParMel19, LinSjo19}.
While the calibration of head-mounted eye trackers nowadays typically needs to be performed \emph{per session} (or even more often), realization of a \emph{once-in-a-lifetime} calibration is clearly an exciting and promising goal for further research and development work. 

\section{Conclusion}
We have presented a high-level overview of the hardware and gaze-estimation pipeline of Pupil Invisible glasses. Employing subject- and population-level metrics, we have shown that Pupil Invisible glasses, without the need for a calibration step, provide gaze prediction, which is robust to headset slippage as well as environmental and person-specific factors. In particular, our data attests that gaze-estimation accuracy in real-world scenarios is on the order of $4^\circ$ of random error.
\begin{center}
	\rule{0.9\linewidth}{0.5pt}
\end{center}
\vspace*{0.2cm}
\appendix
\balance
In this Appendix, we provide a refresher with regard to 2D Gaussian distributions. In particular, we show in what sense they can be conceptualized as shifted ellipses. 

A vector-valued random variable $X=(x_1, x_2)$ is said to have a 2D Gaussian (or normal) distribution if and only if there is vector $\mu\in\mathbb{R}^2$ and a symmetric positive-definite matrix $\mathbf{\Sigma}\in\mathbb{R}^{2\times 2}$, such that its probability density function $p$ can be written as
\begin{equation}\label{eq:density}
p(x;\mu,\mathbf{\Sigma})=\frac{1}{2\pi|\mathbf{\Sigma}|^{1/2}}\rm{exp}\left(-\frac{1}{2}(x-\mu)^T\mathbf{\Sigma}^{-1}(x-\mu)\right),
\end{equation}
where $|\mathbf{\Sigma}|$ is the determinant of $\mathbf{\Sigma}$.
We denote this by \mbox{$X\sim \mathcal{N}(\mu, \mathbf{\Sigma})$}.
The vector $\mu$ is referred to as the mean vector of $X$, since 
\begin{equation}
E[X]=\mu,
\end{equation}
i.e. $\mu$ equals the expected value of $X$. 
The matrix $\mathbf{\Sigma}$ is referred to as the covariance matrix of $X$, since
\begin{equation}
{\rm Cov}[X,X]=E[(X-E[X])(X-E[X])^T]=\mathbf{\Sigma}.
\end{equation}
If the off-diagonal entries of $\mathbf{\Sigma}$ are non-vanishing, then $x_1$ and $x_2$ are correlated. 
Being a symmetric matrix, $\mathbf{\Sigma}$ has three independent entries. 
Note, in contrast to a 1D normal distribution, which is characterized by two parameters (mean and standard deviation), in the 2D case, five parameters are needed (two for $\mu$, three for $\mathbf{\Sigma}$). 
\begin{figure}
	\captionsetup{singlelinecheck=off}
	\centering
	\includegraphics[width=0.94\columnwidth]{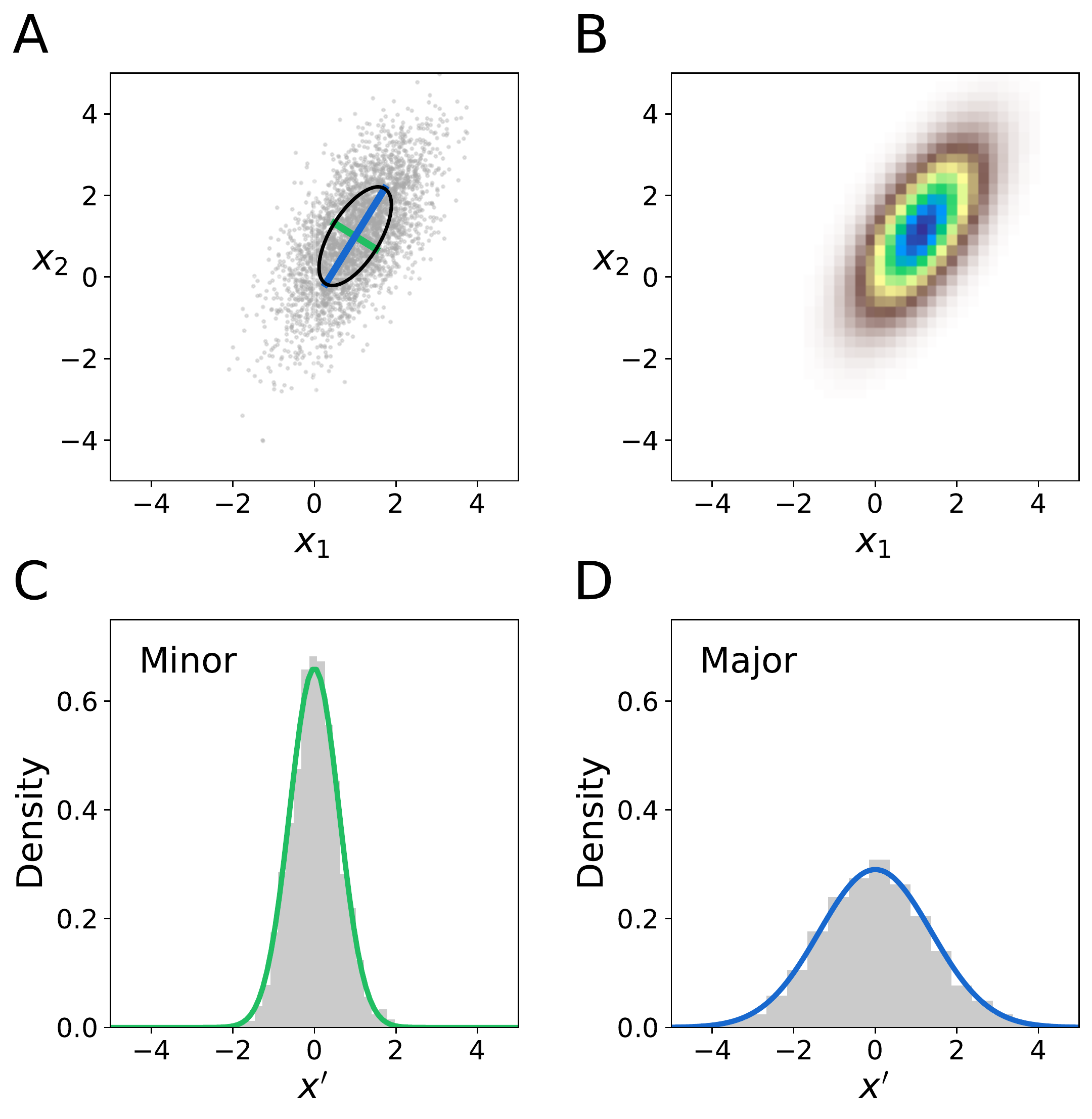}
	\caption{
		\label{fig:gaussian} Geometry of the 2D Gaussian distribution.
		(A) Gray dots are samples drawn from a 2D Gaussian distribution with $\mu=(1.0,1.0)$ and 
		$\mathbf{\Sigma}=(0.7, 0.8; 0.8, 2.5).$ 
		We also show the ellipse with minor (green) and major (blue) axis as defined by $\mathbf{\Sigma}$ and $\mu$. 
		(B) Density of 2D Gaussian distribution in A shown as a heatmap. 
		(C) Measuring the density of samples in A away from $\mu$ along the axis defined by the minor axis, after proper normalization results in a 1D normal distribution (green line) with standard deviation given by $\sigma=\sqrt\lambda_2$. 
		(D) Similar to C, the density along the major direction after proper normalization is a 1D Gaussian (blue line) with $\sigma=\sqrt\lambda_1$.
	}
\end{figure}
The parameters defining a 2D Gaussian distribution can be interpreted in geometric terms. 
Similar to the 1D case, the mean vector $\mu$ specifies the location of the distribution (see Fig.~\ref{fig:gaussian}A). 
As to the covariance $\mathbf{\Sigma}$, first note that isoprobability curves for a 2D Gaussian distribution take the form of ellipses (see Fig.~\ref{fig:gaussian}A), with the directions of the major and minor axis, respectively, coinciding for any two of them.
Their respective directions are related to the direction of eigenspaces of $\mathbf{\Sigma}$.
More specifically, since $\mathbf{\Sigma}$ is symmetric and positive definite, it has two positive real eigenvalues, $\lambda_1$ and $\lambda_2$.
Without loss of generality, we can assume that $\lambda_1\geq \lambda_2$.
Eigenvectors for $\lambda_1$ correspond to the direction of the major axes, eigenvectors for $\lambda_2$ to the direction of the minor axes (in case $\lambda_1=\lambda_2$, any pair of orthogonal eigenvectors will correspond to the directions of major and minor axis). 
Evaluating the density given by Eq.~(\ref{eq:density}) along the minor axis defined by an eigenvector for $\lambda_2$, after proper normalization results in a 1D normal density with standard deviation $\sqrt{\lambda_2}$ (see Fig.~\ref{fig:gaussian}C). 
A corresponding statement holds for $\lambda_1$ and the major axis (see Fig.~\ref{fig:gaussian}D).
In summary, while the covariance matrix $\mathbf{\Sigma}$ encodes the orientation of isoprobability curves (ellipses) and the spread of the density along their major and minor axes, the mean vector $\mu$ determines the center of these ellipses.

\bibliographystyle{plain}
\bibliography{bibliography}

\end{document}